\documentclass{elegantpaper}
\usepackage[T1]{fontenc}    
\usepackage{url}            
\usepackage{amsfonts}       
\usepackage{nicefrac}       
\usepackage{microtype}      
\usepackage{graphicx}
\usepackage{subfigure}
\usepackage{algorithm}
\usepackage{algorithmic}
\usepackage{amssymb}
\usepackage{amsmath}
\usepackage{multirow}
\title{Clustered Reinforcement Learning}

\author{{Xiao Ma, Shen-Yi Zhao and Wu-Jun Li} %
		\\[0.5ex] %
		National Key Laboratory for Novel Software Technology\\
  Collaborative Innovation Center of Novel Software Technology and Industrialization\\
  Department of Computer Science and Technology, Nanjing University, China\\
  \texttt{\{max, zhaosy\}@lamda.nju.edu.cn, liwujun@nju.edu.cn}\\}

\begin{document}
\maketitle
\begin{abstract}

 Exploration strategy design is one of the challenging problems in reinforcement learning~(RL), especially when the environment contains a large state space or sparse rewards. During exploration, the agent tries to discover novel areas or high reward~(quality) areas. In most existing methods, the novelty and quality in the neighboring area of the current state are not well utilized to guide the exploration of the agent. To tackle this problem, we propose a novel RL framework, called \underline{c}lustered \underline{r}einforcement \underline{l}earning~(CRL), for efficient exploration in RL. CRL adopts clustering to divide the collected states into several clusters, based on which a bonus reward reflecting both novelty and quality in the neighboring area~(cluster) of the current state is given to the agent. Experiments on a continuous control task and several \emph{Atari 2600} games show that CRL can outperform other state-of-the-art methods to achieve the best performance in most cases.

\end{abstract}

\section{Introduction}

Reinforcement learning~(RL)~\cite{DBLP:books/lib/SuttonB98} studies how an agent can maximize its cumulative reward in an unknown environment, by learning through exploration and exploiting the collected experience. A key challenge in RL is to balance the relationship between exploration and exploitation. If the agent explores more novel states, it might never find rewards to guide the learning direction. Otherwise, if the agent exploits rewards too intensely, it might converge to suboptimal behaviors and have fewer opportunities to discover more rewards from exploring.

Although reinforcement learning, especially deep RL~(DRL), has recently attracted much attention and achieved significant performance in a variety of applications, such as game playing~\cite{DBLP:journals/nature/MnihKSRVBGRFOPB15,DBLP:journals/nature/SilverHMGSDSAPL16} and robot navigation~\cite{DBLP:conf/icra/ZhangMFLA16}, exploration techniques in RL are far from satisfactory in many cases. Exploration strategy design is still one of the challenging problems in RL, especially when the environment contains a large state space or sparse rewards. Hence, it has become a hot research topic to design exploration strategy, and many exploration methods have been proposed in recent years.

Some heuristic methods for exploration, such as $\epsilon$-greedy~\cite{DBLP:journals/nature/SilverHMGSDSAPL16,DBLP:books/lib/SuttonB98}, uniform sampling~\cite{DBLP:journals/nature/MnihKSRVBGRFOPB15} and i.i.d./correlated Gaussian noise~\cite{DBLP:journals/corr/LillicrapHPHETS15,DBLP:conf/icml/SchulmanLAJM15}, try to directly obtain more different samples~\cite{DBLP:conf/nips/BellemareSOSSM16} during exploration. For hard applications or games, these heuristic methods are insufficient enough and the agent needs exploration techniques that can incorporate meaningful information about the environment.

In recent years, some exploration strategies try to discover novel state areas for exploring. The direct way to measure novelty is using the counts. In~\cite{DBLP:conf/nips/BellemareSOSSM16,DBLP:conf/icml/OstrovskiBOM17}, pseudo-count is estimated from a density model. Hash-based method~\cite{DBLP:conf/nips/TangHFSCDSTA17} records the visits of hash codes as counts. There also exist some approximate ways for computing counts~\cite{DBLP:journals/ml/KearnsS02,DBLP:journals/jmlr/BrafmanT02,DBLP:conf/nips/AuerO06,DBLP:journals/ftml/GhavamzadehMPT15,DBLP:conf/agi/SunGS11,DBLP:conf/icml/KolterN09,DBLP:conf/nips/GuezHSD14}. Besides, the state novelty can also be measured by empowerment~\cite{DBLP:conf/cec/KlyubinPN05}, the agent's belief of environment dynamics~\cite{DBLP:conf/nips/HouthooftCCDSTA16}, prediction error of the system dynamics model~\cite{DBLP:conf/icml/PathakAED17,DBLP:journals/corr/StadieLA15}, prediction by exemplar model~\cite{DBLP:conf/nips/FuCL17}, and the error of predicting features of states~\cite{DBLP:journals/corr/abs-1810-12894}.

All the above methods perform exploration mainly based on the novelty of states, without considering the quality of states. Furthermore, in most existing methods, the novelty and quality in the neighboring area of the current state are not well utilized to guide the exploration of the agent. To tackle this problem, we propose a novel RL framework, called \underline{c}lustered \underline{r}einforcement \underline{l}earning~(CRL), for efficient exploration in RL. The contributions of CRL are briefly outlined as follows:
\begin{itemize}
\item CRL adopts clustering to divide the collected states into several clusters. The states from the same cluster have similar features. Hence, the clustered results in CRL provide a possibility to share meaningful information among different states from the same cluster.
\item CRL proposes a novel bonus reward, which reflects both novelty and quality in the neighboring area of the current state. Here, the neighboring area is defined by the states which share the same cluster with the current state. This bonus reward can guide the agent to perform efficient exploration, by seamlessly integrating novelty and quality of states.
\item Experiments on a continuous control task and several \emph{Atari 2600}~\cite{DBLP:journals/jair/BellemareNVB13} games with sparse rewards show that CRL can outperform other state-of-the-art methods to achieve the best performance in most cases. In particular, on several games known to be hard for heuristic exploration strategies, CRL achieves significant improvement over baselines.
\end{itemize}

The rest content of this paper is organized as follows. Section 2 introduces the related work. Section 3 presents the details of CRL, including the clustering algorithm and clustering-base bonus reward. Section 4 is the experimental result and analysis. Section 5 concludes the paper.

%
\section{Related Work}
\label{Related Work}
In the tabular setting, there is a finite number of state-action pairs that can directly define a decreasing function of the true visitation count as the exploration bonus. MBIE-EB~\cite{DBLP:journals/jcss/StrehlL08} adds the square root of counts of state-action pairs as the bonus reward to the augmented Bellman equation for exploring less visited ones with theoretical guarantee.

In finite MDPs, $E^3$~\cite{DBLP:journals/ml/KearnsS02}, R-Max~\cite{DBLP:journals/jmlr/BrafmanT02} and UCRL~\cite{DBLP:conf/nips/AuerO06} all make use of state-action counts and are activated by the idea of optimism under uncertainty. $E^3$~\cite{DBLP:journals/ml/KearnsS02} determines online to choose an efficient learning policy. R-Max~\cite{DBLP:journals/jmlr/BrafmanT02} assumes the received reward is not in quality area and trains a fictitious model to learn the optimal policy. UCRL~\cite{DBLP:conf/nips/AuerO06} chooses an optimistic policy by using upper confidence bounds. Bayesian RL methods maintain a distribution of belief state as the uncertainty over possible MDPs~\cite{DBLP:journals/ftml/GhavamzadehMPT15,DBLP:conf/agi/SunGS11,DBLP:conf/icml/KolterN09,DBLP:conf/nips/GuezHSD14} and use counts to explore .

In the continuous and high-dimensional space, the number of states is too large to be counted. In~\cite{DBLP:conf/nips/BellemareSOSSM16,DBLP:conf/icml/OstrovskiBOM17}, the exploration bonus reward is designed based on a state pseudo-count quantity, which is estimated from a density model. In the hash-based method~\cite{DBLP:conf/nips/TangHFSCDSTA17}, the hash function encodes states to hash codes and then it explores with the reciprocal of visitation as a reward bonus, which performs well on some hard exploration \emph{Atari 2600} games. Hash-based method is limited by the hash function. Static hashing, using locality-sensitive hashing, is stable but random. Learned hashing, using an autoencoder~(AE) to capture the semantic features, updates during the training time. A related work is~\cite{DBLP:journals/corr/AbelADKS16}, which record the number of cluster center and action pairs which used to select an action from the Gibbs distribution given to a state.

These count-based methods activate the agent by making use of the novelty of states and do not take quality as consideration.
To the best of our knowledge, the novelty and quality in the neighboring area of the current state have not been well utilized to guide the exploration of the agent in existing methods. This motivates the work of this paper. 

\section{Clustered Reinforcement Learning}
\label{gen_inst}
This section presents the details of our proposed RL framework, called \underline{c}lustered \underline{r}einforcement \underline{l}earning~(CRL).

\subsection{Notation}
In this paper, we adopt similar notations as those in~\cite{DBLP:conf/nips/TangHFSCDSTA17}. More specifically, we model the RL problem as a finite-horizon discounted Markov decision process~(MDP), which can be defined by a tuple $\left(\mathcal{S}, \mathcal{A}, \mathcal{P}, r, \rho_{0}, \gamma, T\right)$. Here, $\mathcal{S}\in \mathbb{R}^d$ denotes the state space, $\mathcal{A}\in \mathbb{R}^{m}$ denotes the action space, $\mathcal{P} : \mathcal{S} \times \mathcal{A} \times \mathcal{S} \rightarrow \mathbb{R}$ denotes a transition probability distribution, $r : \mathcal{S} \times \mathcal{A} \rightarrow \mathbb{R}$ denotes a reward function,  $\rho_{0}$ is an initial state distribution, $\gamma \in(0,1]$ is a discount factor, and $T$ denotes the horizon. In this paper, we assume $r\geq0$. For cases with negative rewards, we can transform them to cases without negative rewards. The goal of RL is to maximize $\mathbb{E}_{\pi, \mathcal{P}}\left[\sum_{t=0}^{T} \gamma^{t} r\left(s_{t}, a_{t}\right)\right]$ which is the total expected discounted reward over a policy $\pi$.
\subsection{CRL}
\label{Clustering}

The key idea of CRL is to adopt clustering to divide the collected states into several clusters, and then design a novel cluster-based bonus reward for exploration. The algorithmic framework of CRL is shown in Algorithm~\ref{algorithm}, from which we can find that CRL is actually a general framework. We can get different RL variants by taking different \emph{clustering algorithms} and different \emph{policy updating algorithms}. The details of Algorithm~\ref{algorithm} are presented in the following subsections, including clustering and clustering-based bonus reward.

\begin{algorithm}[t]
\caption{Framework of Clustered Reinforcement Learning~(CRL)}
\label{algorithm}
\begin{algorithmic}
\STATE Initialize the number of clusters $K$, bonus coefficient $\beta$, count coefficient $\eta$
\FOR{each iteration $j$}
\STATE Collect a set of state-action samples $\{(s_i,a_i,r_i)\}^N_{i=1}$ with policy $\pi_j$;
\STATE Cluster the state samples with $f : \mathcal{S} \rightarrow \mathcal{C}$, where $\mathcal{C}=\{C_i\}_{i=1}^{K}$ and $f$ is some \emph{clustering algorithm};
\STATE Compute the cluster assignment for each state $\phi(s_i) = \underset{k}{\operatorname{argmin}}\Vert s_i-C_k\Vert, \forall i: 1\leqslant i\leqslant N, ~ k: 1\leqslant k \leqslant K$;
\STATE Compute sum of rewards $R_k$ using \eqref{R_k} and the number of states $N_k$ using \eqref{N_k}, $\forall k:1\leqslant k \leqslant K$;
\STATE Compute the bonus $b(s_i)$ using \eqref{eq3};
\STATE Update the policy $\pi_j$ using rewards $\left\{ r_i+ b(s_i)\right\}_{i=1}^{N}$ with some \emph{policy updating  algorithm};
\ENDFOR
\end{algorithmic}
\end{algorithm}

 \subsubsection{Clustering}

 Intuitively, both novelty and quality are useful for exploration strategy design. If the agent only cares about novelty, it might explore intensively in some unexplored areas without any reward. If the agent only cares about quality, it might converge to suboptimal behaviors and have low opportunity to discover unexplored areas with higher rewards. Hence, it is better to integrate both novelty and quality into the same exploration strategy.

 We find that clustering can provide the possibility to integrate both novelty and quality together. Intuitively, a cluster of states can be treated as an area. The number of collected states in a cluster reflects the count~(novelty) information of that area. The average reward of the collected states in a cluster reflects the quality of that area. Hence, based on the clustered results, we can design an exploration strategy considering both novelty and quality. Furthermore, the states from the same cluster have similar features, and hence the clustered results provide a possibility to share meaningful information among different states from the same cluster. The details of exploration strategy design based on clustering will be left to the following subsection. Here, we only describe the clustering algorithm.


In CRL, we perform clustering on states. Assume the number of clusters is $K$, and we have collected $N$ state-action samples $\{(s_i,a_i,r_i)\}^N_{i=1}$ with some policy. We need to cluster the collected states $\{s_i\}_{i=1}^{N}$ into $K$ clusters by using some clustering algorithm $f: \mathcal{S} \rightarrow \mathcal{C}$, where $\mathcal{C}=\{C_i\}_{i=1}^{K}$ and $C_i$ is the center of the $i$th cluster. We can use any clustering algorithm in the CRL framework. Although more sophisticated clustering algorithms might be able to achieve better performance, in this paper we just choose k-means algorithm~\cite{DBLP:series/lncs/CoatesN12} for illustration. K-means is one of the simplest clustering algorithms with wide applications. The detail of k-means is omitted here, and readers can find it in most machine learning textbooks.


\subsubsection{Clustering-based Bonus Reward}
\label{Quality}

As stated above, clustering can provide the possibility to integrate both novelty and quality together for exploration. Here, we propose a novel clustering-based bonus reward, based on which many \emph{policy updating algorithms} can be adopted to get an exploration strategy considering both novelty and quality.

Given a state $s_i$, it will be allocated to the nearest cluster by the cluster assignment function $\phi(s_i)=\underset{k}{\operatorname{argmin}}\Vert s_i-C_k\Vert$. Here, $1\leqslant k \leqslant K$ and $\Vert s_i-C_k\Vert$ denotes the distance between $s_i$ and the $k$th cluster center $C_k$. The sum of rewards in the $k$th cluster is denoted as $R_{k}$, which can be computed as follows:
\begin{equation}
R_{k} = \sum_{i=1}^{N} r_{i} \mathbb{I}(\phi(s_{i})=k),
\label{R_k}
\end{equation}
where $\mathbb{I}(\cdot)$ is an indicator function. $R_k$ is also called \emph{cluster reward} of cluster $k$ in this paper. The number of states in the $k$th cluster is denoted as $N_{k}$, which can be computed as follows:
\begin{equation}
N_{k} = \sum_{i=1}^{N} \mathbb{I}(\phi(s_{i})=k).
\label{N_k}
\end{equation}

Intuitively, a larger $N_k$ typically means that the area corresponding to cluster $k$ has more visits~(exploration), which implies the novelty of this area is lower. Hence, the bonus reward should be inversely proportional to $N_k$. The average reward of cluster $k$, denoted as $\frac{R_{k}}{N_{k}}$, can be used to represents the quality of the corresponding area of cluster $k$. Hence, the bonus reward should be proportional to $\frac{R_{k}}{N_{k}}$.

With the above intuition, we propose a clustering-based bonus reward $b: \mathcal{S} \rightarrow \mathbb{R}$ to integrate both novelty and quality of the neighboring area of the current state $s$ , which is defined as follows:
\begin{equation}
b(s)=\beta\frac{\max (\mathbb{I}(\sum\limits_{i=1}^{N}r_i>0)\eta, R_{\phi(s)})}{N_{\phi(s)}},
\label{eq3}
\end{equation}
where $\beta\in \mathbb{R}^{+}$ is the bonus coefficient and $\eta\in \mathbb{R}^{+}$ is the count~(novelty) coefficient. Typically, $\eta$ is set to a very small number. There are two cases:

\begin{align*}
b(s)=\left\{
\begin{array}{cl}
\beta \frac{\max( \eta,R_{\phi(s)})}{N_{\phi(s)}},&\;\text{if}\;\; \sum\limits_{i=1}^{N}r_i>0,\\
0, & \;\text{if}\;\; \sum\limits_{i=1}^{N}r_i = 0.
\end{array} \right.
\end{align*}

Please note that in this paper, we assume $r_i \geq 0$.

In the first case, $\mathbb{I}(\sum\limits_{i=1}^{N}r_i>0)=1$, which means that the current policy can get some rewards in some states. Hence, the states with rewards can share meaningful information among different states from the same cluster. Please note that $\mathbb{I}(\sum\limits_{i=1}^{N}r_i>0)=1$ only means that there exist some clusters with positive reward, and does not mean all clusters have positive reward. It is possible that all states in some clusters have zero rewards. Please note that $\eta$ is typically set to be a very small positive number. In general, as long as there exist one or two states with positive rewards in cluster $\phi(s)$, $R_{\phi(s)}$ will be larger than $\eta$. Hence, if $b(s)$ = $\frac{\beta\eta}{N_{\phi(s)}}$, it is highly possible that all states in cluster $\phi(s)$ have zero reward. Hence, when $R_{\phi(s)} = 0$ which means no rewards have been got for cluster $\phi(s)$, the bonus reward should be determined by the count of the cluster. This is just what our bonus reward function in~(\ref{eq3}) does. From~(\ref{eq3}), larger $N_{\phi(s)}$ will result in smaller bonus reward $b(s)$. This will guide the agent to explore novel areas corresponding to clusters with less visits~(exploration), which is reasonable. When $R_{\phi(s)} >0$, typically $b(s) =\frac{\beta R_{\phi(s)}}{N_{\phi(s)}}$. For two clusters with the same cluster reward, the cluster with smaller number of states~(higher novelty) will be more likely to be explored, which is reasonable. For two clusters with the same number of states, the cluster with higher cluster reward~(higher quality) will be more likely to be explored, which is also reasonable.

In the second case, $\mathbb{I}(\sum\limits_{i=1}^{N}r_i>0)=0$, which means that the policy is unbelievable and sharing information among different states from the same cluster is not a good choice. Furthermore, the states explored by the current policy should not get any extra bonus reward. This is just what our bonus reward function in~(\ref{eq3}) does.

Hence, the clustering-based bonus reward function defined in~(\ref{eq3}) is intuitively reasonable, and it can seamlessly integrate both novelty and quality into the same bonus function. Finally, the agent will adopt $\{(s_i,a_i,r_i+b_i)\}^N_{i=1}$ to update the policy~(perform exploration). Many \emph{policy updating algorithms}, such as trust region policy optimization~(TRPO)~\cite{DBLP:conf/icml/SchulmanLAJM15}, can be adopted. Please note that $r_i+b_i$ is only used for training CRL in Algorithm~\ref{algorithm}. But the performance evaluation~(test) is measured without $b_i$, which can be directly compared with existing RL methods without extra bonus reward.

%
%
%

\section{Experiments}
\label{Experiments}

We use a continuous control task and several \emph{Atari 2600} games to evaluate CRL and baselines. We want to investigate and answer the following research questions:
\begin{itemize}
\item Is the count-based exploration sufficient to excite the agent to achieve the final goal?

\item Can CRL improve performance significantly across different tasks?

\item What is the impact of hyperparameters on the performance?
\end{itemize}

Due to space limitation, the hyperparameter settings are reported in the supplementary material.

\subsection{Experimental Setup}

\subsubsection{Environments}

\begin{figure*}[tb]
\centering
\begin{minipage}[t]{0.45\linewidth}
\centering
\includegraphics[height=0.735\textwidth,width=1\textwidth]{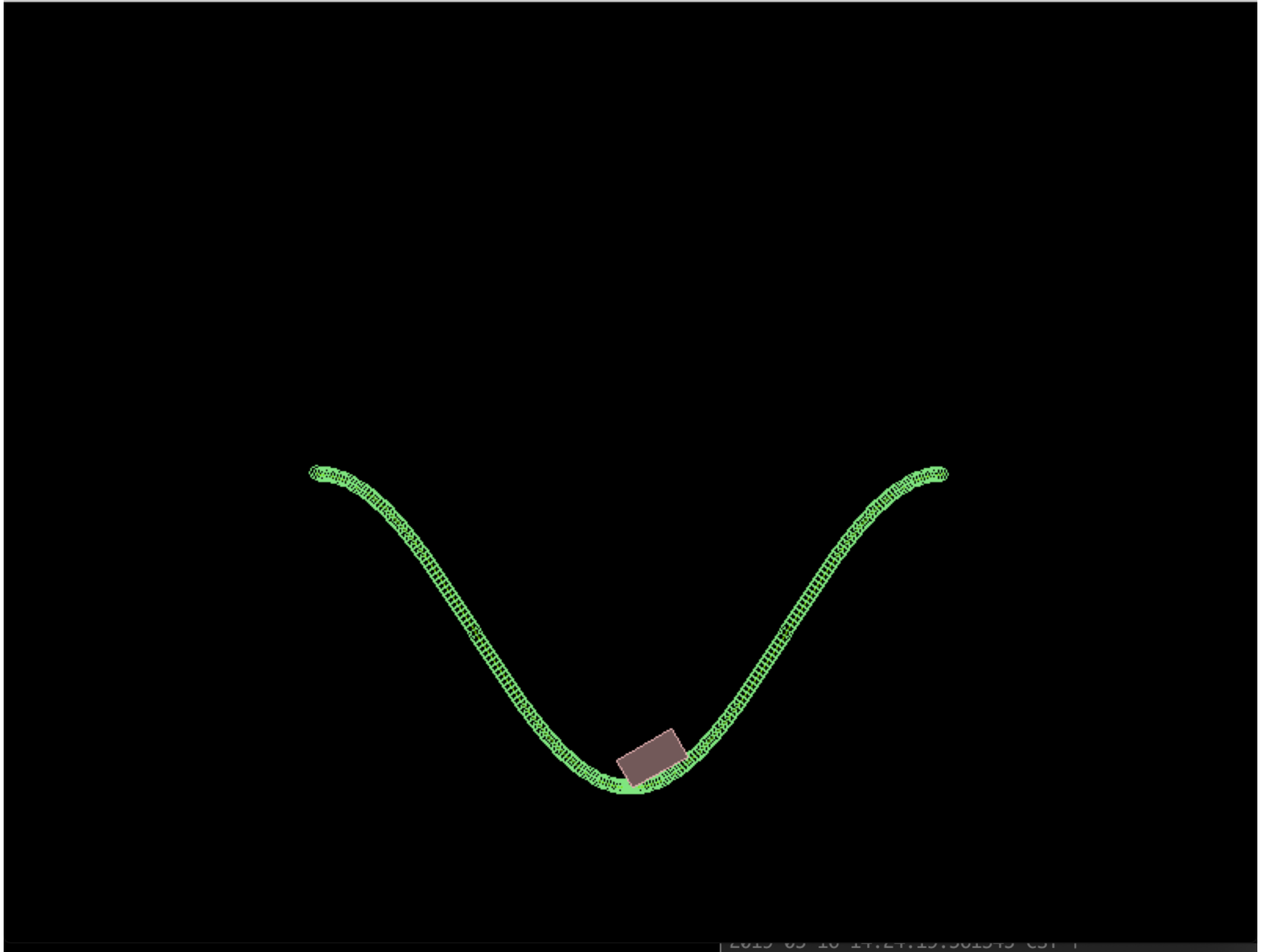}
\\ (a)
\end{minipage}
\hspace{1cm}
\begin{minipage}[t]{0.45\linewidth}
\centering
\includegraphics[width=1\textwidth]{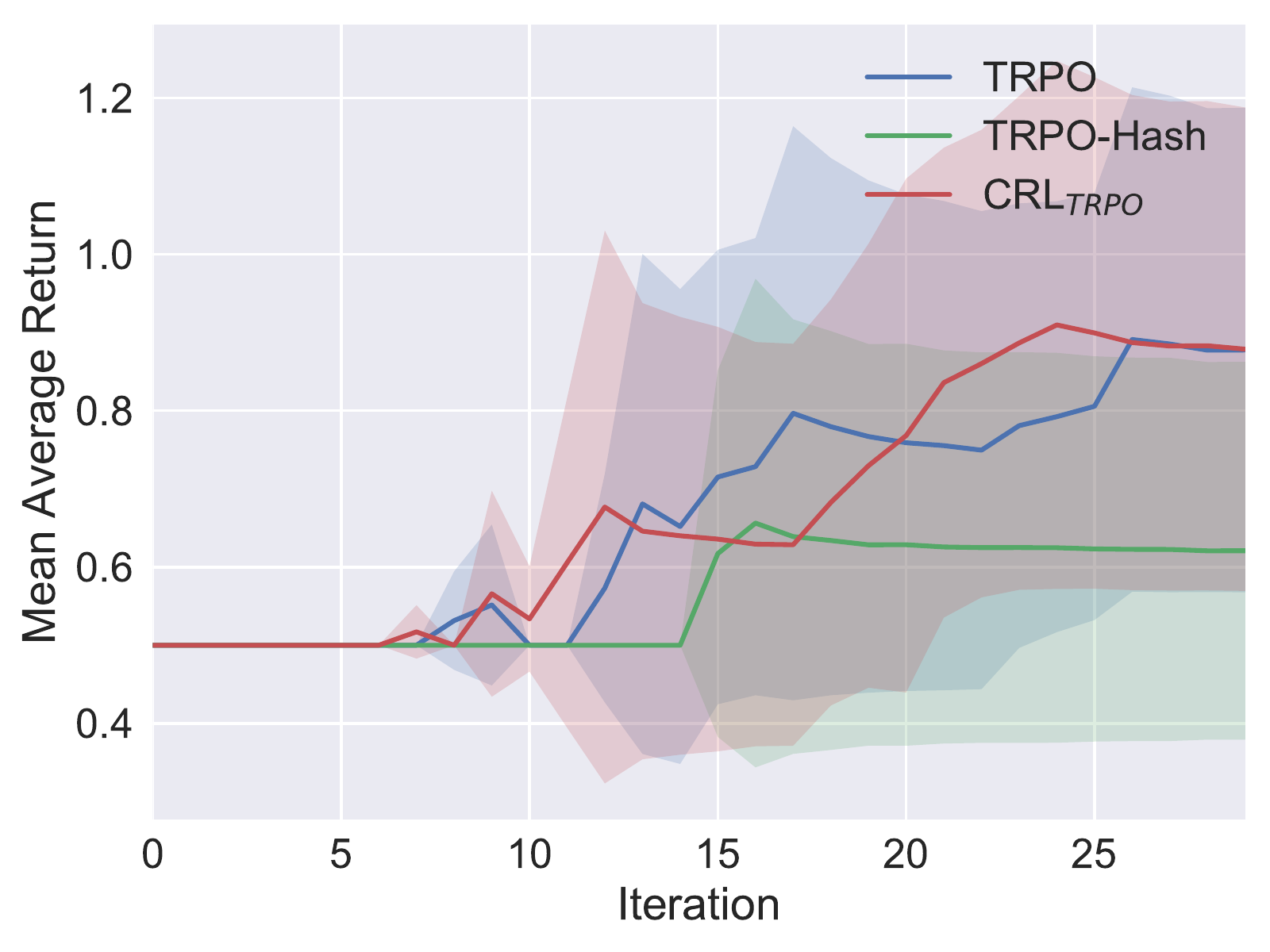}
\\ (b)
\end{minipage}
\caption{\small (a) A snapshot of MountainCar; (b) Mean average return of different algorithms on MountainCar over 5 random seeds. The solid line represents the mean average return and the shaded area represents one standard deviation.}
\label{MuJoCo}
\end{figure*}

\emph{MuJoCo.} The rllab benchmark~\cite{DBLP:conf/icml/DuanCHSA16} consists of various continuous control tasks to test RL algorithms. We design a variant of MountainCar, in which $\mathcal{S} \subseteq \mathbb{R}^{3}, \mathcal{A} \subseteq \mathbb{R}^{1}$. The agent receives a reward of $+1$ when the car escapes the valley from the right side and receives a reward of $+0.001$ at other positions.  One snapshot of this task is shown in Figure~\ref{MuJoCo}~(a).

\begin{figure}[t]
\centering
\subfigure[freeway]{
\label{fig:freeway} 
\includegraphics[width=2.55cm]{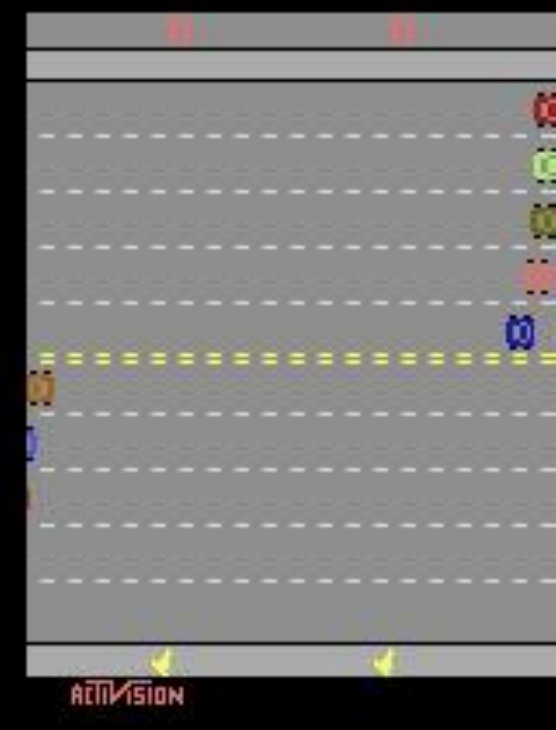}}
\subfigure[frostbite]{
\label{fig:subfig:b} 
\includegraphics[width=2.55cm]{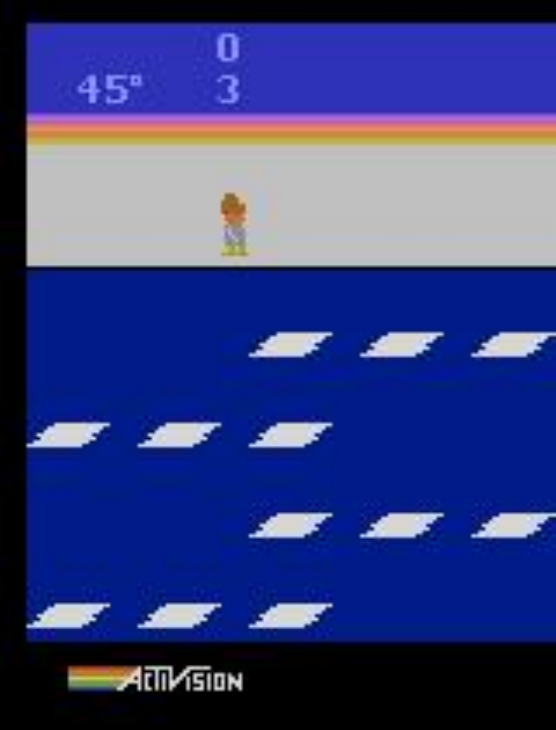}}
\subfigure[gravitar]{
\label{fig:subfig:c} 
\includegraphics[width=2.55cm]{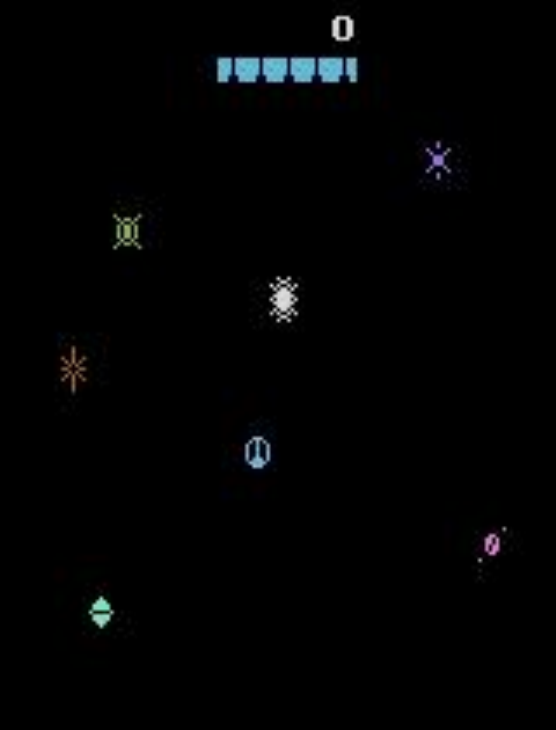}}
\subfigure[solaris]{
\label{fig:a} 
\includegraphics[width=2.55cm]{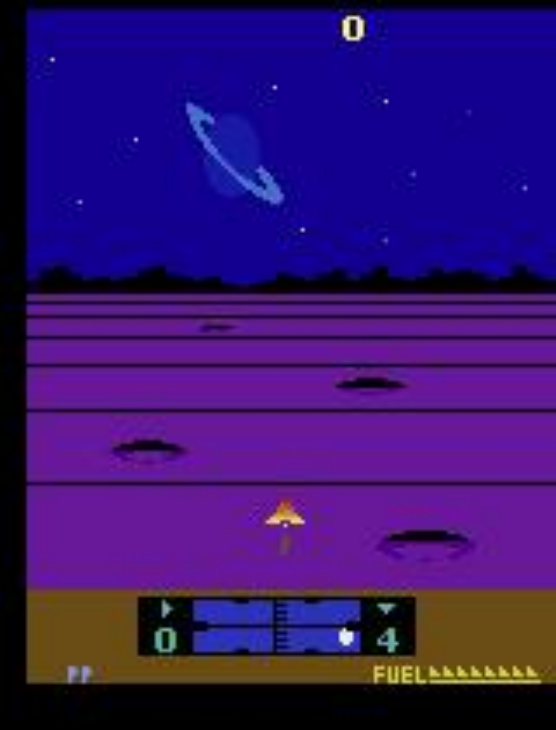}}
\subfigure[venture]{
\label{fig:a} 
\includegraphics[width=2.55cm]{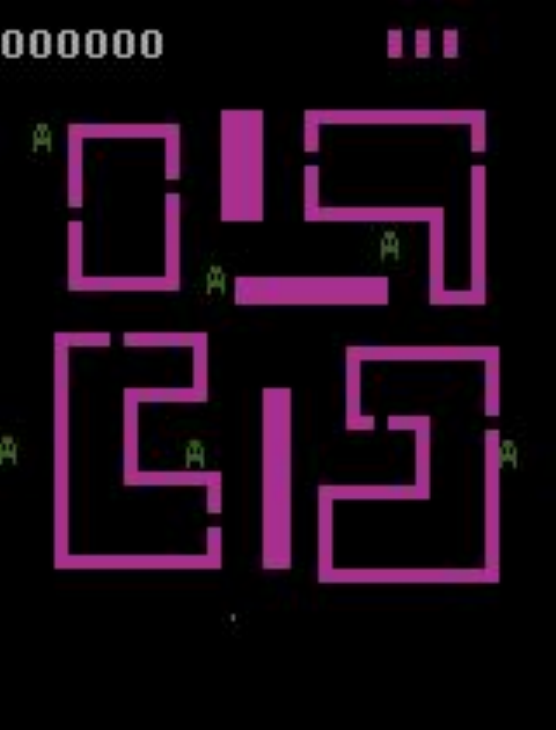}}
\caption{\small Snapshots of five hard exploration \emph{Atari 2600} games.}
\label{figb} 
\vspace{10pt}
\end{figure}

\emph{Arcade Learning Environment.} The Arcade Learning Environment~(ALE)~\cite{DBLP:journals/jair/BellemareNVB13} is an important benchmark for RL because of its high-dimensional state space and wide variety of video games. We select five games\footnote{The Montezuma game evaluated in~\cite{DBLP:conf/nips/TangHFSCDSTA17} is not adopted in this paper for evaluation, because this paper only uses raw pixels which are not enough for learning effective policy on Montezuma game for most methods including CRL and other baselines. We can use advanced feature to learn effective policy, but this is not the focus of this paper.} featuring long horizons and still requiring significant exploration: Freeway, Frostbite, Gravitar, Solaris and Venture. Figure~\ref{figb} shows a snapshot for each game. For example, in Freeway, the agent need to go through the road, avoid the traffic and get the reward across the street. These games are classified into hard exploration category, according to the taxonomy in~\cite{DBLP:conf/nips/BellemareSOSSM16}.
\subsubsection{Baselines}
CRL is a general framework which can adopt many different \emph{policy updating~(optimization) algorithms} to get different variants. In this paper, we only adopt trust region policy optimization~(TRPO)~\cite{DBLP:conf/icml/SchulmanLAJM15} as the policy updating algorithm for CRL, and leave other variants of CRL for future work. We will denote our method as $\mbox{CRL}_{TRPO}$ in the following content. The baselines for comparison include TRPO and TRPO-Hash~\cite{DBLP:conf/nips/TangHFSCDSTA17}, which are also TRPO-based methods and have achieved state-of-the-art performance in many tasks.

\emph{TRPO}~\cite{DBLP:conf/icml/SchulmanLAJM15} is a classic policy gradient method, which uses trust region to guarantee stable improvement of policy and can handle both discrete and continuous action space. Furthermore, this method is not too sensitive to hyperparameters. TRPO adopts a Gaussian control noise as a heuristic exploration strategy.

\emph{TRPO-Hash}~\cite{DBLP:conf/nips/TangHFSCDSTA17} is a hash-based method, which is a generalization of classic count-based method for high-dimensional and continuous state spaces. The main idea is to use locality-sensitive hashing~(LSH)~\cite{DBLP:conf/focs/AndoniI06} to encode continuous and high-dimensional data into hash codes, like $\{-1, 1\}^h$. Here, $h$ is the length of hash codes. TRPO-Hash has several variants in~\cite{DBLP:conf/nips/TangHFSCDSTA17}. For fair comparison, we choose SimHash~\cite{DBLP:conf/stoc/Charikar02} as the hash function and pixels as inputs for TRPO-Hash in this paper, because our CRL also adopts pixels rather than advanced features as inputs. TRPO-Hash is trained by using the code provided by its authors.

\subsection{Disadvantage of Count-based Exploration}
\label{countbad}
TRPO-Hash tries to help the agent explore more novel states and hope that it can achieve better performance than TRPO. But it might go through all states until reaching the goal state, which is the disadvantage of count~(novelty) based exploration. Here, we use MountainCar to show this disadvantage. Figure~\ref{MuJoCo}~(b) shows the results of TRPO, TRPO-Hash and $\mbox{CRL}_{TRPO}$ in MountainCar. We find that TRPO-Hash is slower than TRPO and $\mbox{CRL}_{TRPO}$ on finding out the goal state because the curve of TRPO-Hash ascends until the middle of training. At the end of training, TRPO-Hash fails to reach the goal, but the other methods can achieve the goal. Our method, $\mbox{CRL}_{TRPO}$ is the first to visit the goal and learn to achieve the goal. The reason why TRPO-Hash fails is that the novelty of states diverts the agent's attention. The worst case is that the agent collects all states until it finds the goal. This disadvantage of count-based methods might become more serious in the high-dimensional state space since it is impossible to go through all states in the high-dimensional state space. Therefore, strategies with only count-based exploration are insufficient.

\subsection{Performance on \emph{Atari 2600}}
\label{perform2600}

For video games which typically have high-dimensional and complex state space, advanced features like those extracted by auto-encoder~(AE) and variational auto-encoder~(VAE)~\cite{DBLP:journals/corr/KingmaW13,DBLP:conf/icml/RezendeMW14} can be used for performance improvement. But this is not the focus of this paper. Hence, we simply use raw pixels as inputs for our method and other baselines. Hence, the comparison is fair.

For the five games of \emph{Atari 2600}, the agent is trained for 500 iterations in all experiments with each iteration consisting of 0.4M frames. During each iteration, although the previous four frames are taken into account by the policy and baseline, clustering is performed on the latest frames and counting also pay attention to last frames. The performance is evaluated over 5 random seeds. The seeds for evaluation are the same for TRPO, TRPO-Hash and $\mbox{CRL}_{TRPO}$.

\begin{figure}
\centering
\subfigure[freeway]{
\label{fig:freeway} 
\includegraphics[width=4.5cm,height=4.5cm]{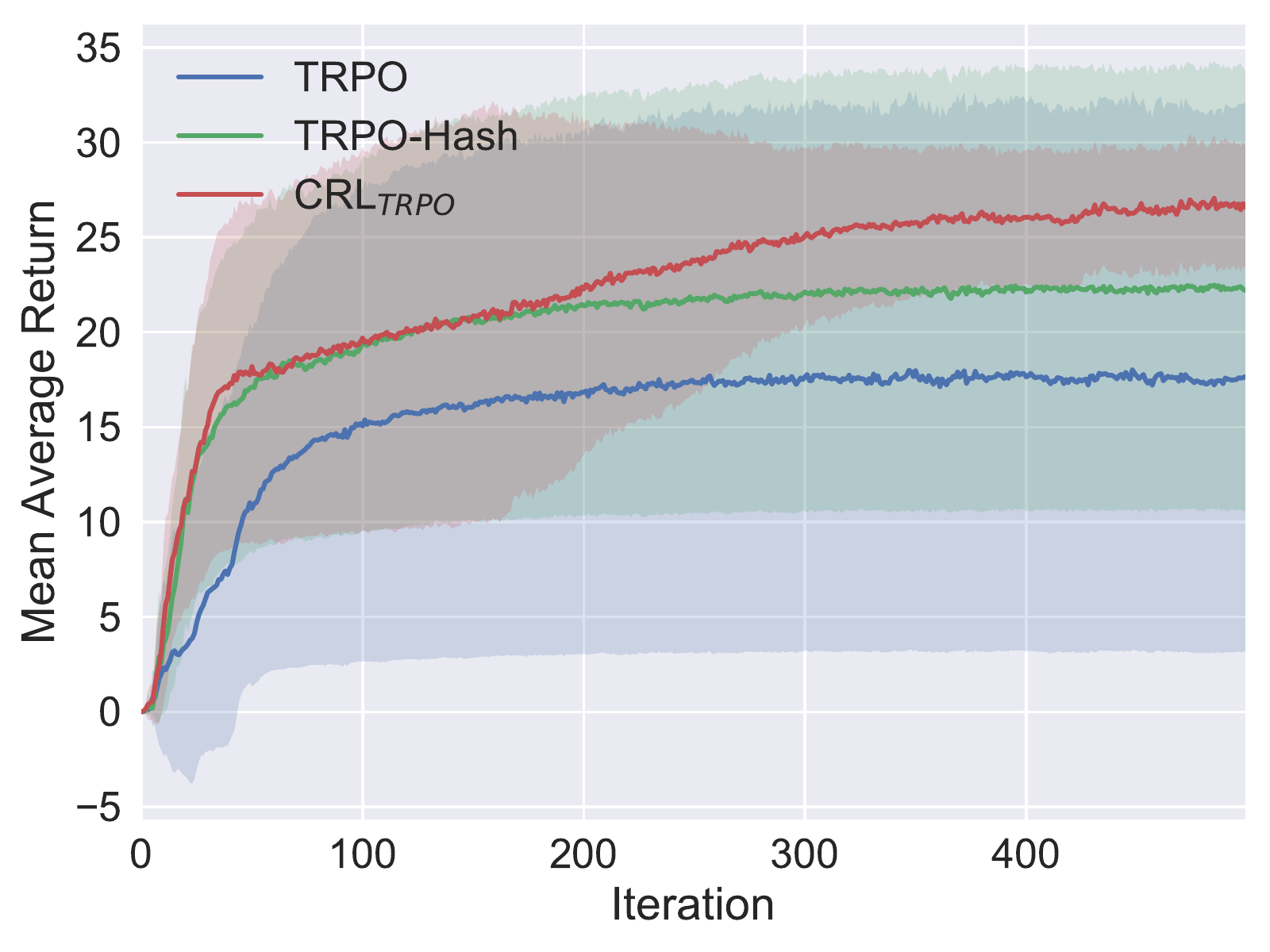}}
\subfigure[frostbite]{
\label{fig:subfig:b} 
\includegraphics[width=4.5cm,height=4.5cm]{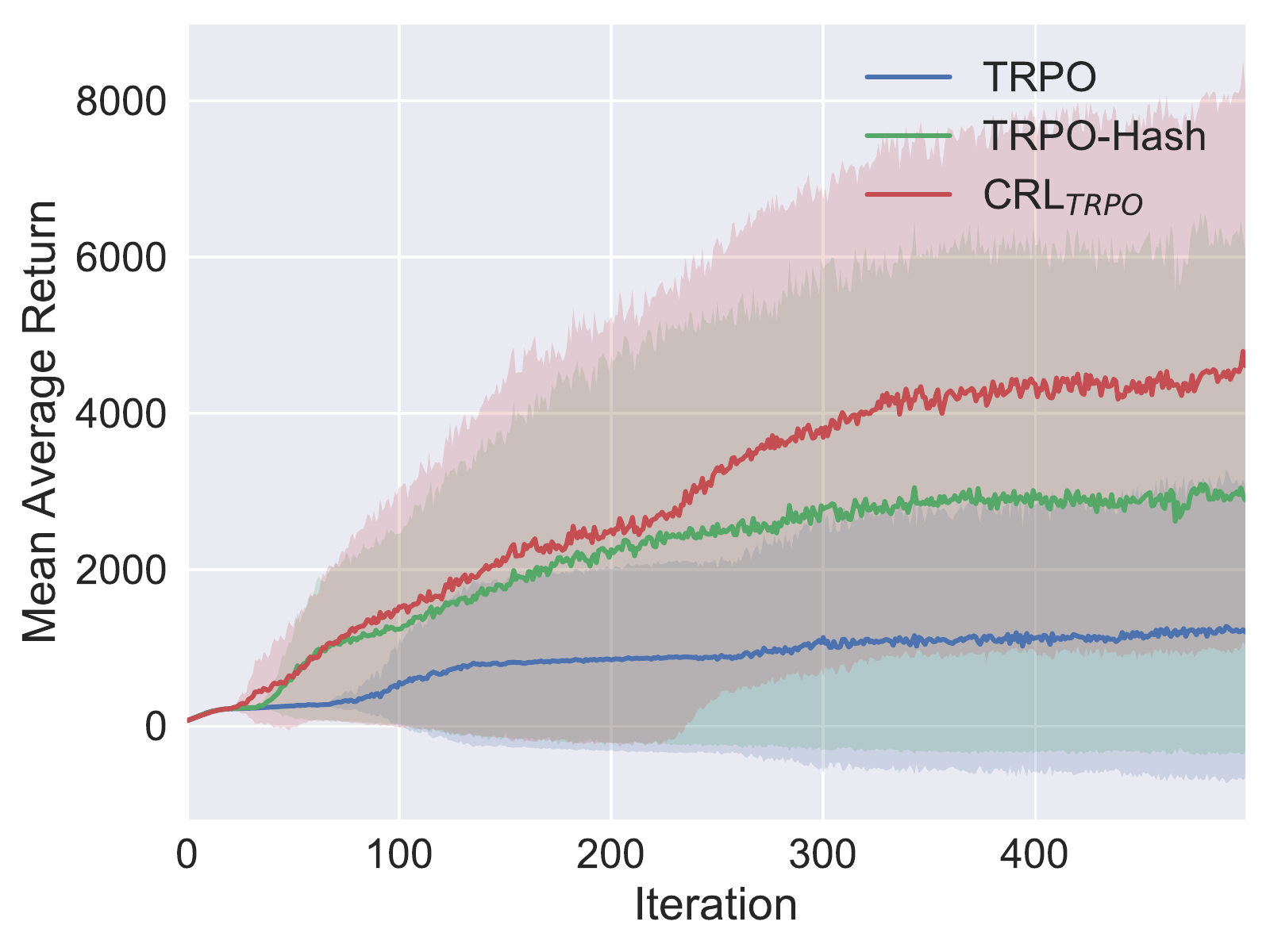}}
\subfigure[gravitar]{
\label{fig:subfig:c} 
\includegraphics[width=4.5cm,height=4.5cm]{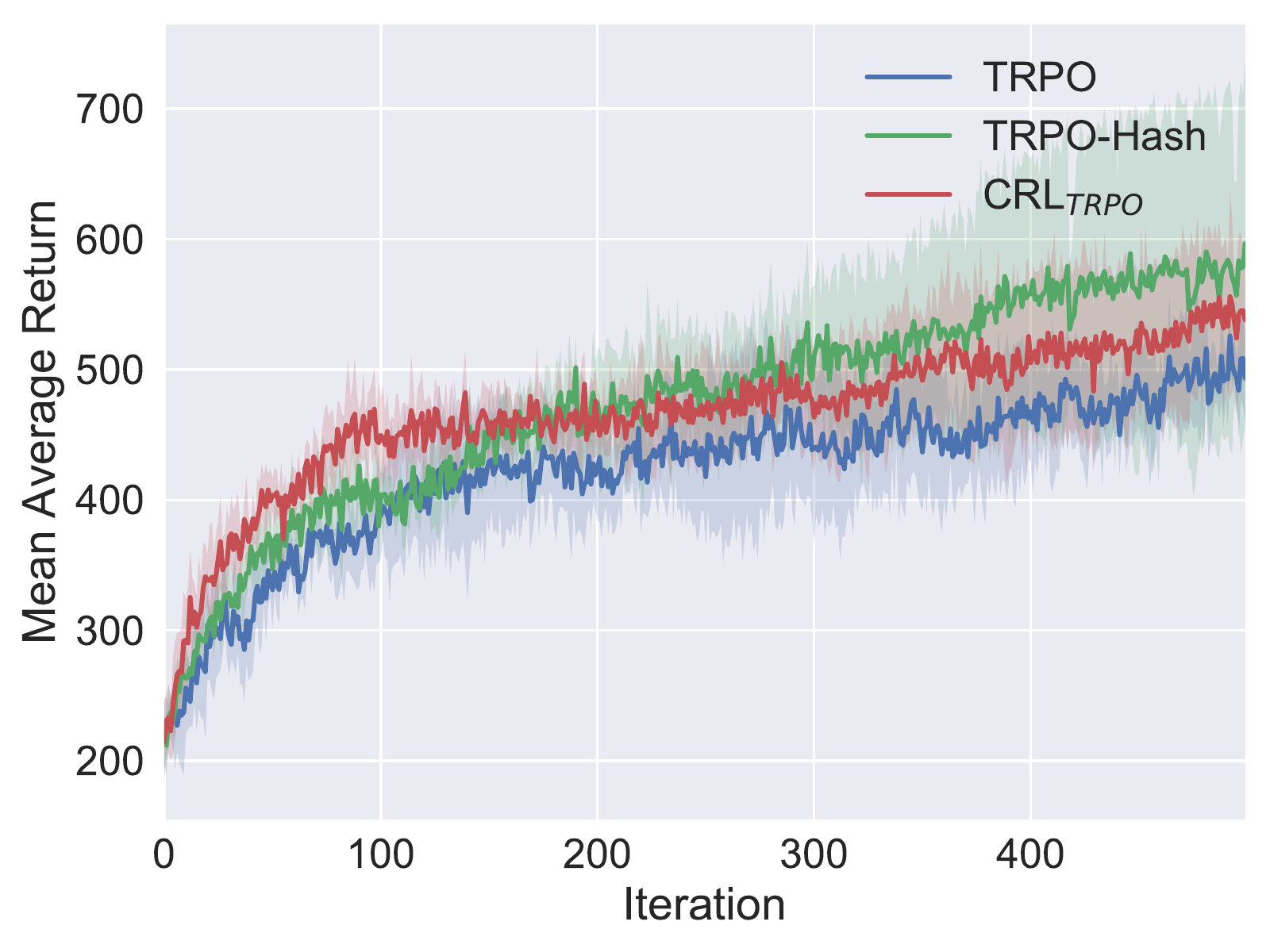}}
\subfigure[solaris]{
\label{fig:a} 
\includegraphics[width=4.5cm,height=4.5cm]{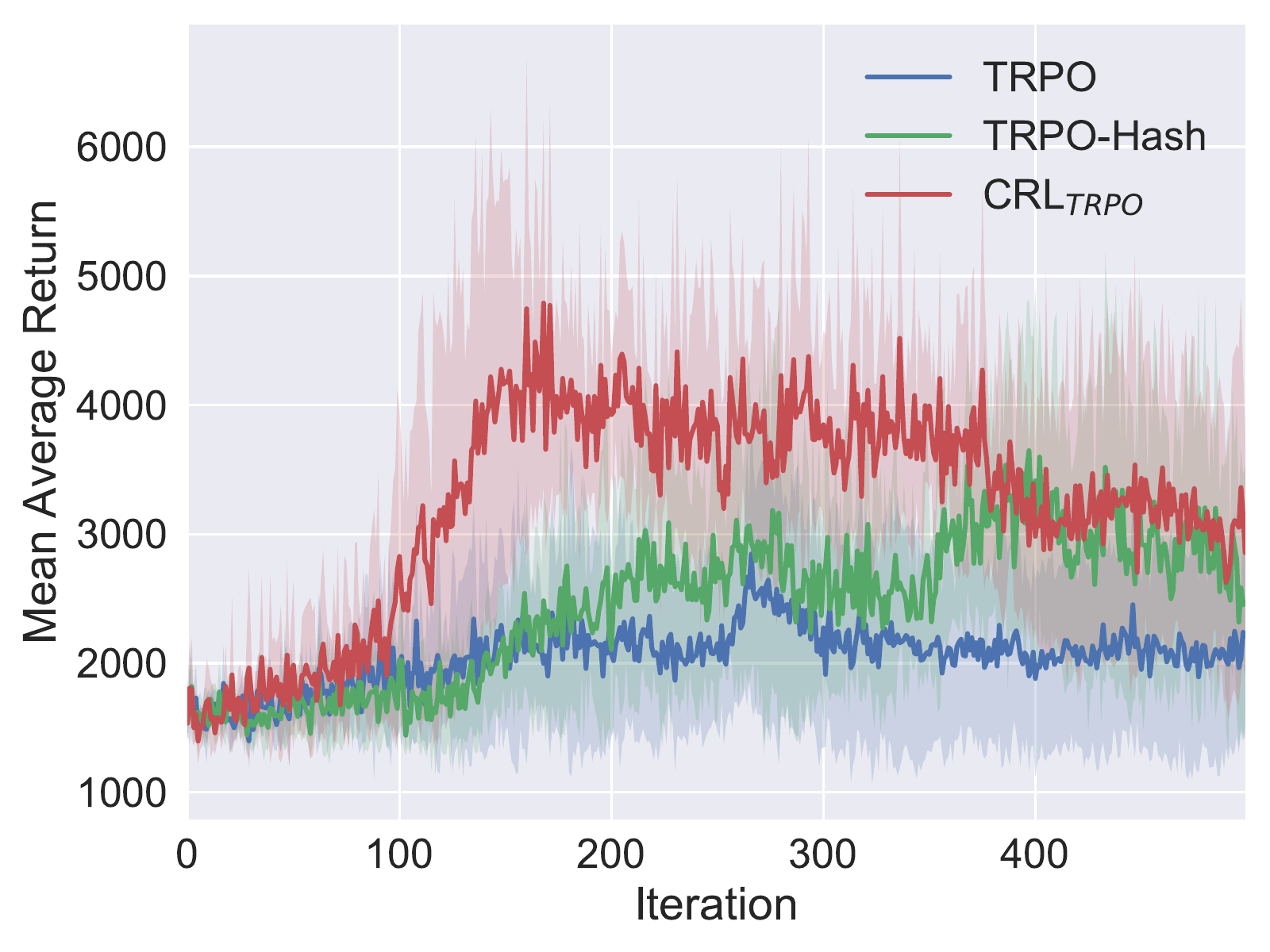}}
\subfigure[venture]{
\label{fig:a} 
\includegraphics[width=4.5cm,height=4.5cm]{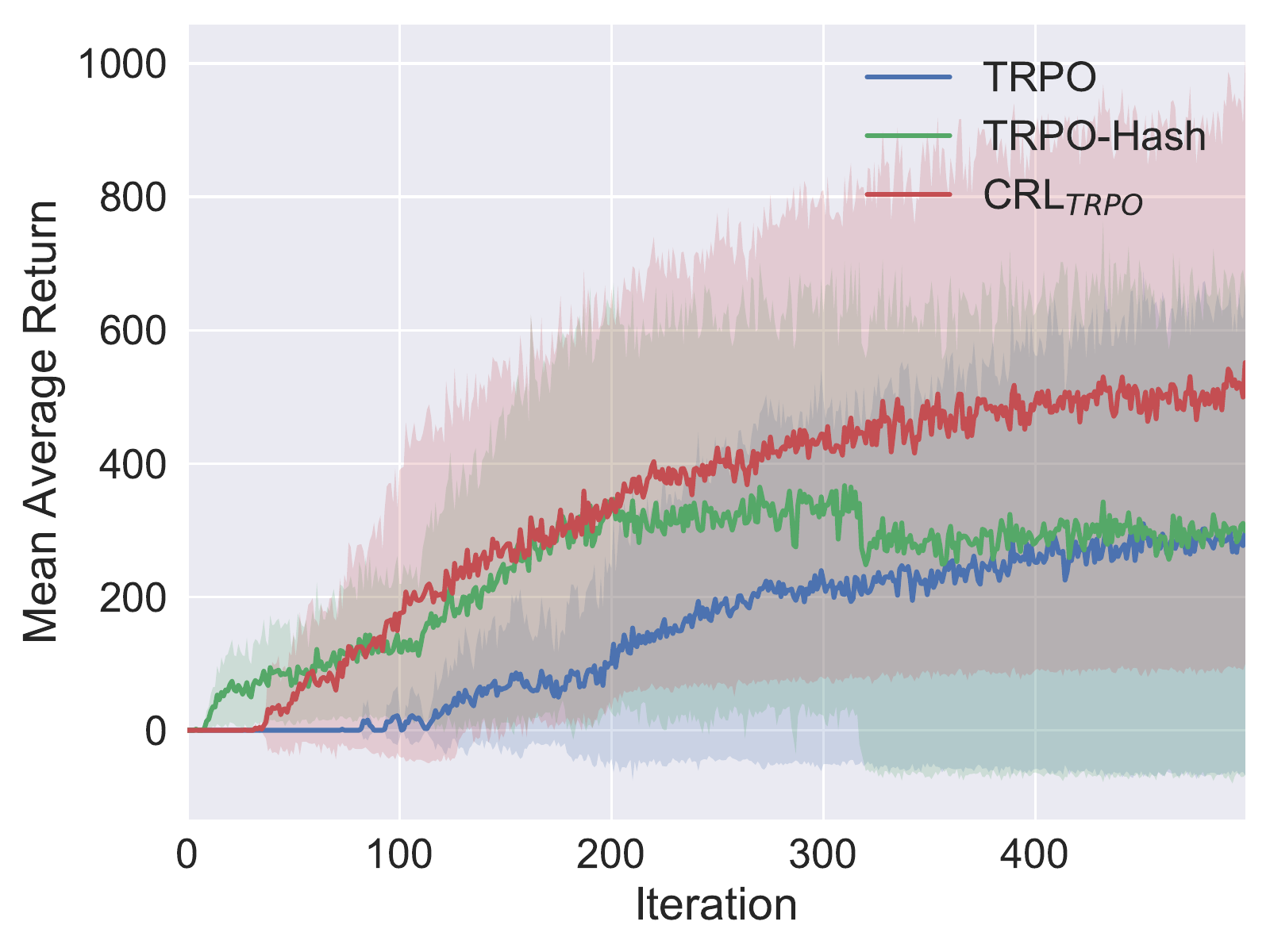}}
\caption{\small Mean average return of different algorithms on \emph{Atari 2600} over 5 random seeds. The solid line represents the mean average return and the shaded area represents one standard deviation.}
\label{AtariResult} 
\end{figure}

We show the training curves in Figure~\ref{AtariResult} and summarize all results in Table~\ref{Tab01}. Please note that TRPO and TRPO-Hash are trained with the code provided by the authors of TRPO-Hash. All hyper-parameters are reported in the supplementary material. We also compare our results to double-DQN~\cite{DBLP:conf/aaai/HasseltGS16}, dueling network~\cite{DBLP:conf/icml/WangSHHLF16}, A3C+~\cite{DBLP:conf/nips/BellemareSOSSM16}, double DQN with pseudo-count~\cite{DBLP:conf/nips/BellemareSOSSM16}, the results of which are from~\cite{DBLP:conf/nips/TangHFSCDSTA17}.


\begin{table}[t]
  \caption{Average total reward after training for 50M time steps. }
  \label{Tab01}
  \centering
  \begin{tabular}{llllll}
    \toprule
    \cmidrule(r){1-2}
                           & Freeway & Frostbite & Gravitar & Solaris & Venture\\
    \midrule
    TRPO                   &  17.55   &  1229.66   &  500.33   &2110.22   &283.48\\
    TRPO-Hash     &  22.29   &  2954.10   & \textbf{577.47}    &   2619.32    &299.61\\
    $\mbox{CRL}_{TRPO}$     &  \textbf{26.68}   & \textbf{4558.52}  &   541.72  &  \textbf{2976.23}  &\textbf{523.79} \\
    \midrule
    Double-DQN             & 33.3    &   1683    & 412      & 3068    & 98.0 \\
    Dueling network        &0.0      &   4672    & 588      & 2251    & 497 \\
    A3C+                   &  27.3   &   507     & 246      & 2175    & 0\\
    pseudo-count           &  29.2   &   1450    & -        & -        & 369\\
    \bottomrule
  \end{tabular}
\end{table}

$\mbox{CRL}_{TRPO}$ achieves significant improvement over TRPO and TRPO-Hash on Freeway, Frostbite, Solaris and Venture. Especially on Frostbite, $\mbox{CRL}_{TRPO}$ achieves more than $250\%$ improvement compared with TRPO and more than $50\%$ improvement compared with TRPO-Hash. On Venture, $\mbox{CRL}_{TRPO}$ achieves more than $70\%$ improvement compared with TRPO and TRPO-Hash. Furthermore, $\mbox{CRL}_{TRPO}$ can outperform all other methods in most cases. Please note that DQN-based methods reuse off-policy experience, which is an advantage over TRPO. Hence, DQN-based methods have better performance than TRPO. But our $\mbox{CRL}_{TRPO}$ can still outperform DQN-based methods in most cases. It is illustrated that the novelty and quality in the neighboring area of states give the on-policy agent an experience buffer like off-policy.

\subsection{Hyperparameter Effect}
\label{param}
We use Venture of \emph{Atari 2600} to study the performance sensitivity to hyperparameters, including $K$ in k-means, $\beta$ and $\eta$ in bonus reward.


We choose different $K$ from $\{8,12,16,20\}$ on Venture to illustrate the effect of $K$ when $\eta=10^{-4},\beta=0.01$. Larger $K$
will divide the state space more precisely but the statistic of average reward $\frac{R_{k}}{N_{k}}$ might become less meaningful. Smaller $K$ will mix information from different areas, which might be too coarse for exploration. The results on Venture are summarized in Table~\ref{Tab05} with fixed $\beta$ and $\eta$. On Venture, the scores are roughly concave in $K$, peaking at around $K=12$. We can find that the performance is not too sensitive to $K$ in a relatively large range.
\begin{table}[ht]
\centering
\caption{Effect of the number~($K$) of clusters on Venture}
\label{Tab05}
\begin{tabular}{cccccc}
\toprule
    $K$      & 8   & 12  & 16   &  20   \\
\midrule
Venture   & 347.86    & 663.32    &  523.79    &   451.94\\
\bottomrule
\end{tabular}
\end{table}



We choose $\beta$ from $\{0.01,0.1\}$ and $\eta$ from $\{0, 10^{-4},10^{-3},10^{-2},10^{-1}\}$. The results are shown in Table~\ref{Tab03}. The value of $\beta \times \eta$ decides the level about pure exploration of novelty. Therefore, by fixing $\beta \times \eta$ as $10^{-6}, 10^{-5}, 10^{-4}$, the performance in $\beta=0.01$ are better than $\beta=0.1$ because large $\beta$ causes the bonus rewards to overwhelm the true rewards. By fixing $\beta = 0.01$, $\eta$ decides the degree of pure exploration about the novel states. Larger $\eta$ means that more novel states will be explored. The scores are roughly concave, peaking at around $\eta=0.0001$, which shows that count-based exploration is insufficient.
\begin{table}[ht]
\centering
\caption{Effect of hyperparameter $\beta$ and $\eta$ on Venture, where the number in bracket is $\beta\times\eta$. }
\label{Tab03}
\begin{tabular}{ccccccc}
\toprule
\multirow{2}{*}{$\beta$} & \multicolumn{5}{c}{$\eta$} \\
\cmidrule(r){2-6}
&  0      &  0.0001  &   0.001  &  0.01      &  0.1     \\
\midrule
0.01  &292.39    & 523.79 (10e-7)    & 512.74 (10e-6)    & 279.84 (10e-5)    &  182.04 (10e-4) \\
0.1& 218.44   & 113.12 (10e-6)    & 101.95 (10e-5)   & 81.70(10e-4)    & 88.51 (10e-3)   \\
\bottomrule
\end{tabular}
\end{table}

\section{Conclusion}
In this paper, we propose a novel RL framework, called \underline{c}lustered \underline{r}einforcement \underline{l}earning~(CRL), for efficient exploration. By using clustering, CRL provides a general framework to adopt both novelty and quality in the neighboring area of the current state for exploration. Experiments on a continuous control task and several hard exploration \emph{Atari 2600} games show that CRL can outperform other state-of-the-art methods to achieve the best performance in most cases.

\small
\bibliography{ref}
\bibliographystyle{abbrv}

\newpage
\appendix
\section{Hyperparameter Settings}
\subsection{Hyperparameter setting in MuJoCo}
In MuJoCo, the hyperparameter setting is shown as Table~\ref{Hyper01}.
\begin{table}[H]
  \caption{Hyperparameter setting in MuJoCo}
  \label{Hyper01}
  \centering
  \begin{tabular}{c|ccc}
    \toprule
                          \ &TRPO & TRPO-Hash & $\mbox{CRL}_{TRPO}$ \\  \hline
    TRPO batchsize         &\multicolumn{3}{c}{5000} \\
    TRPO stepsize          & \multicolumn{3}{c}{0.01} \\
    Discount factor       &   \multicolumn{3}{c}{0.99} \\
    Policy hidden units   &\multicolumn{3}{c}{(32, 32)}\\
    Baseline function     & \multicolumn{3}{c}{Linear} \\
    Iteration             &  \multicolumn{3}{c}{30}\\
    Max length of path    &  \multicolumn{3}{c}{500} \\
     \hline
    Bonus coefficient     &{-}&  {0.01}   &  1\\
    Others                &   {-}   & {Simhash dimension: 32}  &  $\#$cluster centers: 16\\
                  &  {-}  & {-}     &$\eta =0.0001$\\
    \bottomrule
  \end{tabular}
\end{table}

\subsection{Hyperparameter settings in \emph{Atari 2600}}

The hyperparameter settings of results in Figure~\ref{AtariResult} and~Table~\ref{Tab01} is shown as Table~\ref{Hyper02} and Table~\ref{Hyper03}.

In TRPO-Hash,~\cite{DBLP:conf/nips/TangHFSCDSTA17} chooses 16,64,128,256,512 as the SimHash dimension. When SimHash dimension is 16, it only has 65536 distinct hash codes. When SimHash dimension is 64, it has more than $10^{19}$ hash codes and the agent only receive $5\times 10^7$ states during the training time. Therefore, we choose 64 as the SimHash dimension, which is sufficient. The hyperparameters settings about exploration of TRPO-Hash and TRPO-Clustering are shown in~Table~\ref{Hyper03}.

We choose smaller $\eta$ for Venture because Venture belongs to hard exploration category with sparse rewards. As the analyze in Section~\ref{param}, large $\beta$ might mislead the agent to novel but low-quality area because the bonus is decided by the novelty part severely. Therefore, we choose $\eta=0.0001$ for Venture.
\begin{table}[H]
  \caption{Hyperparameter setting in \emph{Atari 2600}}
  \label{Hyper02}
  \centering
  \begin{tabular}{l|c}
    \toprule
                           & TRPO, TRPO-Hash, $\mbox{CRL}_{TRPO}$ \\ \hline

    TRPO batchsize         &  100K  \\

    TRPO stepsize          &0.001 \\

    Discount factor       & 0.99  \\
    Iteration             & 500 \\
    Max length of path    &  4500 \\
     \hline
    Policy structure      & {16 conv filters of size $8 \times 8$, stride 4}\\
                          & {32 conv filters of size $4 \times 4$, stride 2}\\
                          & {fully-connect layer with 256 units}\\
                          & {linear transform and softmax to output action probabilities}\\
      \hline
    Input preprocessing  & {grayscale; downsampled to $42 \times 42$; each pixel rescaled to $[-1, 1]$}  \\
                         & { 4 previous frames are concatenated to form the input state}\\
    \bottomrule
  \end{tabular}
\end{table}

\begin{table}[H]
  \caption{Hyperparameter setting of Exploration in the Table~\ref{AtariResult}}
  \label{Hyper03}
  \centering
  \begin{tabular}{c|c|c}
    \toprule
                           & TRPO-Hash & $\mbox{CRL}_{TRPO}$ \\
 \hline
   bonus coeffcient $\beta$ &  {0.01}  & {0.01}\\
    \hline

      others & SimHash dimension: 64& number of cluster centers: 16 \\
      &&venture: $\eta = 0.0001$\\
      &&others: $\eta = 0.1$\\
    \bottomrule
  \end{tabular}
\end{table}

\section{Hyperparameter sensitivity in Frostbite}
Frostbite is easier than Venture because of dense rewards, although Frostbite is one of games in the hard exploration category. In Frostbite, we achieve more than $250\%$ of the baseline (TRPO) and more than $50\%$ of TRPO-Hash method score. And due to space limitation, we show the hyperparameter effects on this Section.
\subsection{The hyperparameter $K$ in k-means}
Similar to Venture, we choose different $K$ from $\{8,12,16,20\}$ when $\eta=10^{-4},\beta=0.01$. When $K=20$, it is too large to make the information in cluster centers useless. When $K=8$, the performance has significant improvement. It is illustrated that the choice of $K$ needs to balance the relationship between segmentation granularity and statistics difference.
\begin{table}[H]
\centering
\caption{The effect of the number of cluster center $K$ on Frostbite}
\label{Tab05}
\begin{tabular}{cccccc}
\toprule
         & 8   & 12  & 16   &  20   \\
\midrule
Frostbite &  6275.06    &    2249.02 &  4526.88   &  1346.49  \\
\bottomrule
\end{tabular}
\end{table}

\subsection{The hyperparameter of bonus}
When $K=16$, we choose $\beta$ from $\{0.01,0.1\}$ and $\eta$ from $\{0, 10^{-4},10^{-3},10^{-2},10^{-1}\}$. We fix the value of $\beta \times \eta$ and $\beta =0.01$ performs better than $\beta=0.1$ in most cases. When $\beta$ is fixed as $0.01$, the performances are better than TRPO. The performances have no significantly trend because this game has dense rewards. Therefore, the bonus is affected by the novelty slightly.
\begin{table}[H]
\centering
\caption{The effect of hyperparameter $\beta$ and $\eta$ on Frostbite, where the number in brackets is $\beta \times \eta$. }
\label{Hyper04}
\begin{tabular}{ccccccc}
\toprule
\multirow{2}{*}{$\beta$} & \multicolumn{5}{c}{$\eta$} \\
\cmidrule(r){2-6}
&  0      &  0.0001  &   0.001  &  0.01      &  0.1     \\
\midrule
0.01  &3292.63    &4526.88 (10e-7)    & 2719.07 (10e-6)    & 3691.03 (10e-5)    &  4558.52 (10e-4) \\
0.1& 2835.28   & 766.28 (10e-6)    & 4125.28 (10e-5)   & 2350.22 (10e-4)    & 497.64 (10e-3)   \\
\bottomrule
\end{tabular}
\end{table}

\end{document}